\documentclass[letterpaper, 10 pt, conference]{ieeeconf}  

\IEEEoverridecommandlockouts                              

\usepackage[utf8]{inputenc}
\usepackage{authblk}
\usepackage{microtype}
\usepackage{amsmath} 
\usepackage{flushend}
\usepackage{balance}
\usepackage{bbding}
\usepackage{booktabs}
\title{\LARGE 
Sim-to-Real Transfer of Vision-Language Navigation in\\Continuous Environments Using an Ackermann-Steered Mobile Robot
}

\author{
    \begin{minipage}{0.3\linewidth}
        \centering
        Chalindu Abeywansa\\
        \vspace{-2ex}
        {\small Department of Electronic and Telecommunication Engineering,
        University of Moratuwa,\\
        Colombo, Sri Lanka.\\
        chalindunisal1998@gmail.com}\\
        \vspace{1.5ex}
        Seniru Dissanayake\\
        {\small Department of Electronic and Telecommunication Engineering,
        University of Moratuwa,\\
        Colombo, Sri Lanka.\\
        senirudissanayake@gmail.com}
    \end{minipage}
    \begin{minipage}{0.3\linewidth}
        \centering
        Sahan Gunasekara\\
        \vspace{-2ex}
        {\small Department of Electronic and Telecommunication Engineering,
        University of Moratuwa,\\
        Colombo, Sri Lanka.\\
        skgunasekara99@gmail.com}\\
        \vspace{1.5ex}
        Ranga Rodrigo\\
        {\small Department of Electronic and Telecommunication Engineering,
        University of Moratuwa,\\
        Colombo, Sri Lanka.\\
        ranga@uom.lk}
    \end{minipage}
    \begin{minipage}{0.3\linewidth}
        \centering
        Devindi De Silva\\
        \vspace{-2ex}
        {\small Department of Electronic and Telecommunication Engineering,
        University of Moratuwa,\\ 
        Colombo, Sri Lanka.\\
        devindikds@gmail.com} \\
        \vspace{1.5ex}
        Peshala Jayasekara \\
        {\small Department of Electronic and Telecommunication Engineering,
        University of Moratuwa,\\
        Colombo, Sri Lanka.\\
        peshala@uom.lk}
    \end{minipage}
}

\begin{document}
\maketitle
\thispagestyle{empty}
\pagestyle{empty}

\begin{abstract}

Vision-Language Navigation (VLN) enables robots to navigate through environments using natural language instructions, making human-robot interaction intuitive. 
Traditional VLN models often rely on navigation graphs, 360-degree views, and perfect localization which pose significant challenges when adapting these models to real-world settings. This work addresses these limitations by performing a simulation-to-real domain shift of a VLN approach that operates in continuous environments without requiring navigation graphs or panoramic views. The proposed system integrates vision-language models that align visual inputs and linguistic instructions within a shared embedding space, facilitating natural language-driven navigation. 
We employ a Cross-Modal Attention (CMA) based architecture trained on an existing dataset in a simulated environment and fine-tune it using real-world data collected from a custom-built Ackermann-steered robot equipped with a camera and a LiDAR sensor.
By utilising linear photometric adjustments and fine-tuning on a limited number of episodes, our model successfully adapts to real-world environments, achieving effective navigation while running offline on dedicated hardware. Experimental results, evaluated using Success weighted by Path Length (SPL) and Normalized Dynamic Time Warping (nDTW) metrics, demonstrate the robustness and adaptability of our approach. 

\vspace{0.2cm}
\end{abstract}

\section{INTRODUCTION}
\label{sec:intro}
Vision-Language Navigation (VLN), where robots navigate through an environment by following natural language instructions, has garnered significant attention due to its intuitive approach to robotic control. This emerging field leverages the ability of vision-language models to map visual data (from images captured by the robot) and language instructions into a shared embedding space, enabling the prediction of actions within a specified action space.

Most early VLN studies use simulated environments, which were often discrete~\cite{li2023kerm,chen2022think, anderson2018vision, zhu2021soon, qi2020reverie}. These implementations frequently rely on: (i) oracle navigation, which is capable of accurately traversing multiple meters in the presence of obstacles; (ii) known topology, where agents operate on a fixed network of traversable nodes (graph); (iii) ideal 360-degree panoramic RGB imagery and depth information; and (iv) perfect localization, where agents are given their precise location and heading at all times. These assumptions present a major challenge when transferring VLN models trained in simulated environments to real-world continuous environments.

Transferring a VLN agent from a simulated environment to the real world (sim-to-real) introduces additional complexities \cite{anderson2021sim}. 
Visual domain gaps can exist between the simulated training environment and the real world due to differences in the camera quality, dynamic range, and lighting conditions. 
Viewpoint variations, even slight ones, can degrade the agent's performance.
In simulations, the VLN agent is trained with a high-level discrete action space, yet real robots operate in a continuous low-level motor control space.
While the sim-to-real work by Anderson \emph{et al.}~\cite{anderson2021sim} is notable for being one of the first such implementations, it relies on SLAM \cite{DurrantWhyte2006SimultaneousLA} for navigation in physical world.

To bring VLN significantly closer to real-world scenarios, Krantz \emph{et al.}~\cite{krantz2020beyond}  introduced VLN-CE, which removes the need for discrete graph assumptions while adopting a more generalized approach for navigation in continuous environments. It uses a 3D continuous environment simulator (Habitat) \cite{savva2019habitat} for the VLN task. Nevertheless, as a simulation-based work, it cannot be directly applied to an embodied real-world implementation due to the aforementioned sim-to-real challenges.

In this work, we present a real-world VLN framework leveraging VLN-CE while addressing sim-to-real complications. Our model encodes the image and depth information using two ResNet50 networks and encodes natural language instructions using a Bi-LSTM encoder. Then, through cross-modal attention, we predict an action among a discrete action space. We train our model using the VLN-CE \cite{anderson2018vision} dataset in simulated environment.
Then, through linear photometric adjustments and fine-tuning, we enhance the model performance in real-world settings.
We used a custom Ackermann-steered mobile robot to implement the sim-to-real VLN system utilizing Robot Operating System (ROS) \cite{Quigley2009ROSAO} framework. 
We validate the efficacy of our approach through experiments demonstrating the robot's ability to interpret and follow language instructions across diverse real-world settings. 

The contributions of this work are as follows:
\begin{itemize}
   \item We introduce a VLN sim-to-real architecture without navigation graphs and 360-degree views.
  \item We demonstrate successful navigation in real-world indoor environments using an Ackermann-steered mobile robot with minimal fine-tuning in sim-to-real transfer.
  To the best of our knowledge, our work is the first to attempt VLN using an Ackermann-steered robot.
  \item We provide a ROS implementation for offline VLN system deployment, enabling practical applications.
\end{itemize}


\section{RELATED WORK}

The key developments in VLN can be explored, highlighting the evolution of methods from discrete action spaces to continuous environments, while identifying existing challenges and research gaps.

\subsection{Vision-Language Navigation}
The concept of VLN was first introduced by Anderson \emph{et al.} \cite{anderson2018vision}, and their work laid the foundation for the VLN task, where the agent perceives its surroundings through RGB images and navigates based on high-level language commands. 
This pioneering effort utilized the Room-to-Room (R2R) {\cite{anderson2018vision}} dataset, a benchmark for training and evaluating VLN models in simulated environments. 
Subsequent research in VLN {\cite{li2023kerm,chen2022think,vln-pano2real}} has largely focused on enhancing the model's ability to interpret language and visual inputs more effectively. Early models, such as those based on recurrent neural networks (RNNs) and attention mechanisms, demonstrated the feasibility of combining visual and linguistic modalities for navigation tasks {\cite{krantz2020beyond,krantz2021waypoint}}. However, these models were often constrained to discrete environments and action spaces, limiting their application to pre-defined environments with fixed viewpoints and requiring navigation graphs for path planning.
\subsection{Navigation Graphs }
Models operating in discrete action spaces rely on predefined navigation graphs that map out possible routes within the environment {\cite{chen2021topological, daniuvsis2021topological}}. These navigation graphs consist of nodes (representing specific locations) and edges (representing navigable paths between nodes), facilitating structured movement from one point to another. This approach faces significant limitations when applied to real-world scenarios.
Moreover, the reliance on 360-degree views and exact localization is often unachievable in real-world applications, where visual input may be obstructed or incomplete due to environmental factors.

\subsection{Sim-to-Real Transfer}
Sim-to-real transfer has been a crucial focus in VLN to bridge the gap between performance in simulated environments and real-world applications. 
Early works in VLN achieved notable success by training agents in high-fidelity simulated environments with controlled settings, predefined maps, and panoramic images \cite{chen2022think, li2023kerm, anderson2018vision,  hermann2020learning}. 
These simulated environments allowed researchers to evaluate navigation tasks using structured visual data and idealized conditions, but transitioning these systems to the real world introduced substantial challenges. Visual domain gaps can exist between the simulated environment and the real world due to differences in the camera quality, dynamic range, and lighting conditions.
One of the main difficulties is the sim-to-real gap, which encompasses discrepancies in visual quality, and the sensor fidelity of real-world environments. 
Anderson \emph{et al.} \cite{anderson2021sim} addressed these challenges by employing domain randomization to handle visual differences and using a subgoal prediction model to translate high-level actions from simulation into the continuous, real-world action space required by physical robots. Their work demonstrated that sim-to-real transfer is feasible with the support of pre-mapped environments and annotated navigation graphs, but performance degraded significantly when these aids were removed, underscoring the limitations of sim-to-real strategies in previously unseen settings.
LM-Nav~{\cite{shah2023lm}} leverages pre-exploration and large pretrained models to enhance navigation in complex environments, relying heavily on external processing and prior knowledge. 
In contrast, our implementation does not depend on precomputed maps, allowing the robot to operate autonomously in unknown environments with limited processing power. 

\subsection{Vision-Language Navigation in Continuous Environments}
The transition from discrete, graph-based navigation to continuous environments has been a significant step forward in the development of VLN systems. VLN in Continuous Environments (VLN-CE) by Krantz \emph{et al.} \cite{krantz2020beyond, krantz2021waypoint}, comes first in that transition.
VLN-CE \cite{krantz2020beyond} focuses on enabling agents to operate in more realistic, unbounded settings where actions are not restricted to a set of predefined paths. The action space of VLN-CE results in significantly longer-horizon tasks, with an average of 55.88 steps compared to the 4-6 steps typical in discrete environments (R2R). While recent research enhances VLN in continuous environments, it still relies on the use of pre-computed navigation graphs \cite{an2024etpnav, irshad2021hierarchical, an2023bevbert} and 360-degree panoramic visuals \cite{krantz2021waypoint}, and remains limited to simulation-only implementations. 



\section{METHODOLOGY}

\begin{figure*}[thpb]
  \includegraphics[width=1\linewidth]{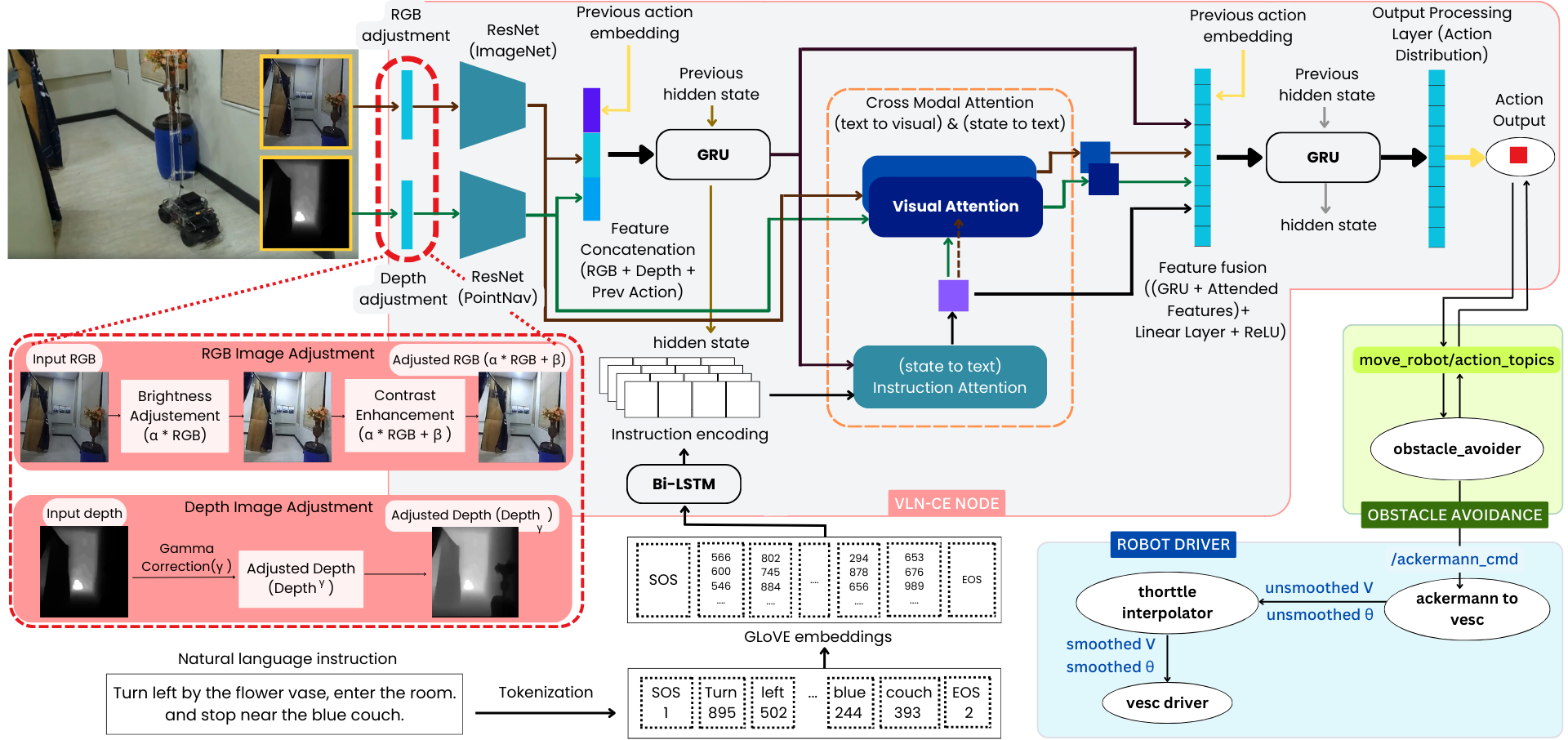}
  \caption {Overall system architecture: 
  RGB and depth features are generated using two ResNet50 networks, while natural language instructions are encoded with a Bi-LSTM. First GRU keeps track of the visual observations and its output is used to generate the instruction attention, and serves as an input to the second GRU. The output from the cross-modal attention, which integrates the instruction and visual features, is also passed into the second GRU. The last linear layer predicts the robot's next action, which is fed to the ROS action handler.}
    \label{fig:The system architecture}
\end{figure*}

Our goal is to build an embodied VLN implementation for unseen environments that carries out navigation tasks using only the egocentric view of the robot and natural language instruction without having the restriction of a predefined navigation graph. We propose a cohesive framework to achieve this task by leveraging cross-modal attention across a natural language model and computer vision models to predict a discrete action within a predefined action space. Our framework also includes the safe execution of this predicted action through a ROS action handler integrated with a fail-safe algorithm for obstacle avoidance. We deploy our system on a custom Ackermann-steered robot platform for testing and evaluation. The following subsections describe these methods in detail. Fig.~\ref{fig:The system architecture} shows the overall system architecture.

\subsection{Generalized System Architecture}

The architecture processes multimodal inputs, including RGB images, depth data, and natural language instructions, to predict navigation actions in continuous environments. 
RGB and depth data captured through camera sensors undergo photometric adjustments, and are encoded using dual ResNet50~\cite{he2016deep} networks pretrained on the ImageNet~\cite{deng2009imagenet} and PointNav~\cite{wijmans2019dd} datasets.
Precomputed instruction embeddings (GLoVE embeddings~{\cite{pennington2014glove}}) are encoded using a Bi-LSTM~\cite{graves2005framewise}, which captures context from both sides of the instruction for accurate language understanding in a memory-efficient manner, enabling deployment on resource-constrained standalone devices. 
Then, these multimodal inputs are processed using a cross-modal attention-based architecture to generate navigation actions, enabling the robot to move through the environment as per the instructions.



\subsubsection{Cross-Modal Attention (CMA) Model}
We adopt a state-of-the-art CMA model architecture~\cite{krantz2020beyond} that effectively implements key techniques common to vision-and-language tasks, such as cross-modal attention and spatial visual reasoning, both essential for language-guided visual navigation.
As illustrated in Fig.~\ref{fig:The system architecture}, the image embeddings concatenated together with the previous action embedding is fed into a Gated Recurrent Unit (GRU)\cite{cho2014learning} to track visual observations. Next, the CMA generates instruction, visual, and depth attention, which are subsequently used by the second GRU to predict the next action embedding.
\subsubsection{Integration with ROS}

ROS is used to manage the execution of predicted navigation actions, acting as the middleware for communication between the VLN model and the robot's hardware components. 
ROS handles the acquisition of sensory data, motor control, and real-time obstacle avoidance, facilitating smooth and coordinated navigation in real-world environments. We believe this integration will help other researchers to adapt VLN in with their own robots already modeled with ROS. We plan to make our code available.
\subsection{Dataset and Simulation Environment} \label{dataset and simulation environment}
Our model was pre-trained on the VLN-CE simulation dataset, created by Krantz \emph{et al.} \cite{krantz2020beyond}. 
Derived from the R2R dataset, VLN-CE extends the VLN task into continuous 3D environments, moving beyond the fixed navigation graph and panoramic images used in R2R. 
The dataset maps trajectories from the original R2R dataset onto navigable paths to be used with the Matterport3DSimulator (MP3D-Sim) \cite{ramakrishnan2021habitat}, enabling agents to navigate continuously between points. Since the VLN-CE dataset features a higher resolution of actions per trajectory,  it offers a more realistic framework for developing robots capable of real-world, instruction-following navigation. Due to these merits, we have incorporated the VLN-CE dataset (Train set: 10837 episodes, validation set: 1839 episodes, test set: 3408 episodes) for the initial training of our model. This dataset's diversity in spatial layouts, textures, and lighting conditions made it suitable for developing robust navigation capabilities in simulation. The Habitat simulator \cite{savva2019habitat}, configured with the Matterport3D environments, facilitated the training process by providing a controlled environment for simulating real-world navigation scenarios.

\subsection{Training in Simulated Environment}
The initial training phase focused on teaching the model to associate visual observations with corresponding linguistic instructions using a combination of reinforcement learning and supervised learning techniques. In reinforcement learning, the model utilized DAgger (Dataset Aggregation) to improve its performance by collecting trajectories and refining its policy through imitation learning, which helped in learning from the actions taken by an oracle agent.

Then, the model was trained on ground-truth paths, along with the corresponding natural language instructions and a pre-computed path followed via low-level actions until convergence using supervised learning, allowing it to learn the correct associations between visual inputs and language instructions.

\subsection{Robot Platform}

\begin{figure}[t]
  \centering
  \includegraphics[width=0.85\linewidth]{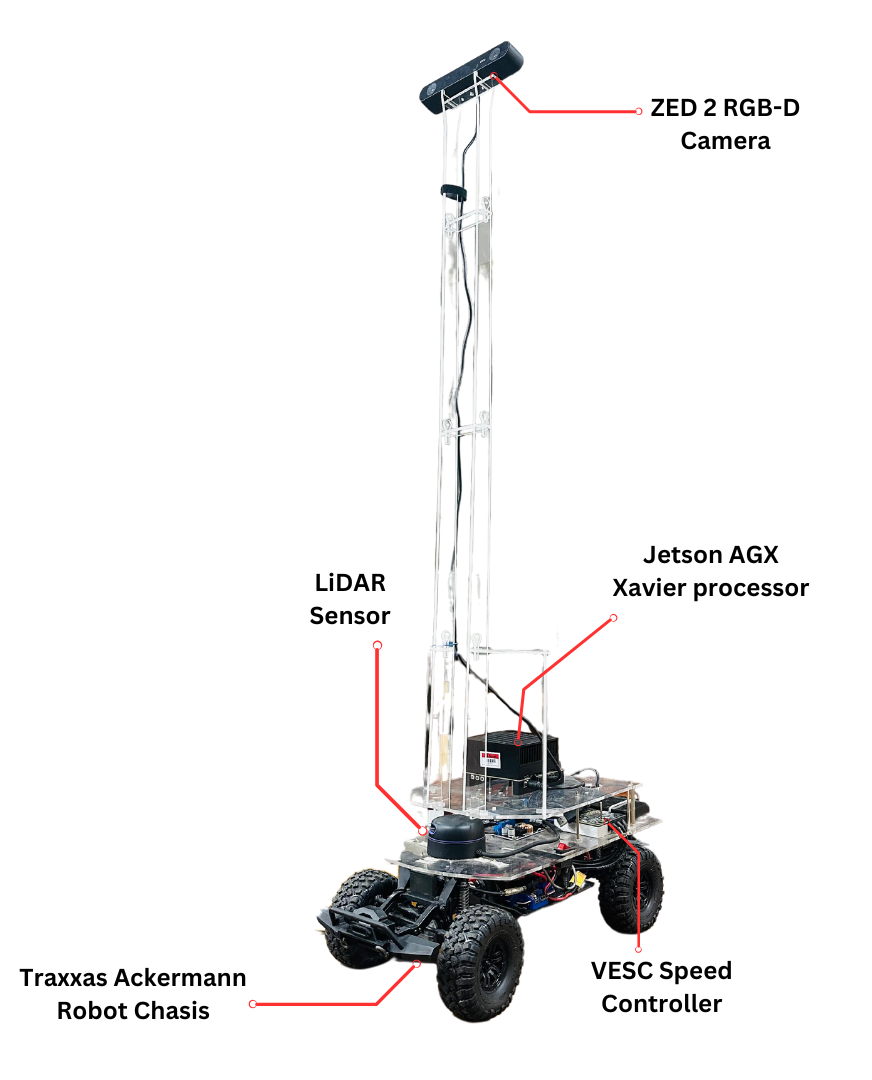}
  \caption{The robot platform with mounted components. The RGB-D camera is elevated to perceive the environment at nearly human height.}
    \label{fig:Robot Platform}
\end{figure}

To evaluate our model's performance in real-world settings, we developed an integrated robotic platform as shown in Fig.~\ref{fig:Robot Platform}. 
The Ackermann-steered robot chassis (width: $0.32$\,m, length: $0.54$\,m) consists of an NVIDIA Jetson AGX Xavier for on-board processing, an RGB-D camera and a LiDAR sensor for environmental sensing. 
The motivation to employ an Ackermann-steered robot in our system is twofold. First, since humans issue language instructions from an egocentric perspective, it is crucial for the robot's camera to have a similar height and viewpoint. The four-wheel robot equipped with an Ackermann-steering mechanism is sturdy for this requirement. It ensures smooth navigation, while its suspension system maintains the stability of the elevated camera. Second, although using an Ackermann-steered platform presents greater challenges compared to standard differential drive robots, we employ it to better align with potential future aspirations, such as autonomous taxis capable of responding to language-based commands.

The ROS-based software architecture integrates the VLN model with the robot’s control system. 
Key components in the ROS system are the VLN node which runs the VLN model, the obstacle avoidance node which continuously monitors the surroundings for any obstacle using the LiDAR to dynamically adjust the steering angle based on the measurement, and a driver node which translates the speed and angle commands to drive the motors and adjust the steering angle.



\textbf{Ackermann-steering} The implementation of an Ackermann-steered robot platform for VLN in indoor environments presents unique challenges, particularly in navigating tight spaces and avoiding obstacles. Unlike differential drive systems commonly used in indoor robotics, Ackermann-steering cannot execute zero-radius turns.
To address these limitations, we developed an obstacle avoidance method suited to the Ackermann-steering geometry. In short, the LiDAR scan data is segmented into front, left, and right regions, with average distances to obstacles calculated for each sector. When obstacles are detected within the robot's vicinity, the system dynamically adjusts both speed and steering angle to maintain a safe trajectory.
To ensure smooth navigation in indoor settings, we implemented an interpolation-based mechanism for motor and servo commands. This approach prevents sudden changes in speed or steering angles, resulting in smooth and controlled movements. As shown in Table~\ref{comparison},  the steering mechanism had less impact on the overall performance of the VLN model.
\subsection{Fine-Tuning the System}
For the fine-tuning process, we compiled a comprehensive dataset by equipping our robot platform with a joystick interface, integrating it with the ROS action handler. The collected data included RGB frames, depth images, textual instructions, and the corresponding actions executed by the human navigator during each episode\cite{dias2020deep}.
Upon receiving each RGB and depth image, we performed photometric adjustments [Sec. ~\ref{sec:datacollection}] to ensure that the real-world input data aligns with the conditions used during training in the simulated environment. 
This preprocessing step helps the model to maintain performance consistency when transitioning from simulated to real-world scenarios, facilitating more accurate visual feature extraction and navigation.


The processed RGB and depth images, along with the corresponding textual instruction, were subsequently fed into the CMA model. 
Specifically, we extract the output of the second GRU with embeddings of $512$ dimensions.
In our fine-tuning phase, we focused solely on refining the last linear layer of the model, responsible for mapping the $512-$dimensional embeddings to one of the four fundamental navigation actions: ``stop", ``move forward", ``turn right", and ``turn left." We calculated the cross entropy loss at the end and backpropagated, updating the weights of the final linear layer.
Due to this targeted approach, despite using a limited dataset for fine-tuning, the model was able to closely match the performance observed in simulation, demonstrating that even minimal real-world data can significantly improve the model's adaptability as shown in our results. 


\section{EXPERIMENTS AND RESULTS} 
This section outlines the experiments used to assess our VLN system's performance in simulated and real-world environments, aiming to validate its capability to interpret natural language instructions and navigate unstructured spaces. We use standard metrics (SPL and nDTW) to evaluate the model and analyze the effectiveness of our training and fine-tuning strategies similar to \cite{li2023kerm, chen2022think, krantz2020beyond, anderson2018evaluation}.
\subsection{Experimental Setup}
\subsubsection{Simulation Environment}
The initial phase of experimentation was conducted in a controlled simulation environment using the Habitat simulator, configured with the VLN-CE dataset (Sec. \ref{dataset and simulation environment}). 
The simulated environment matches real-world scenarios to the extent possible, including the presence of obstacles, doors, and furniture, which the robot must navigate around while following language instructions. Experiments in simulation are conducted through 5247 episodes in the VLN-CE dataset (test and validation set).

\begin{figure}[b]
      \centering
      \includegraphics[width=1.0\linewidth]{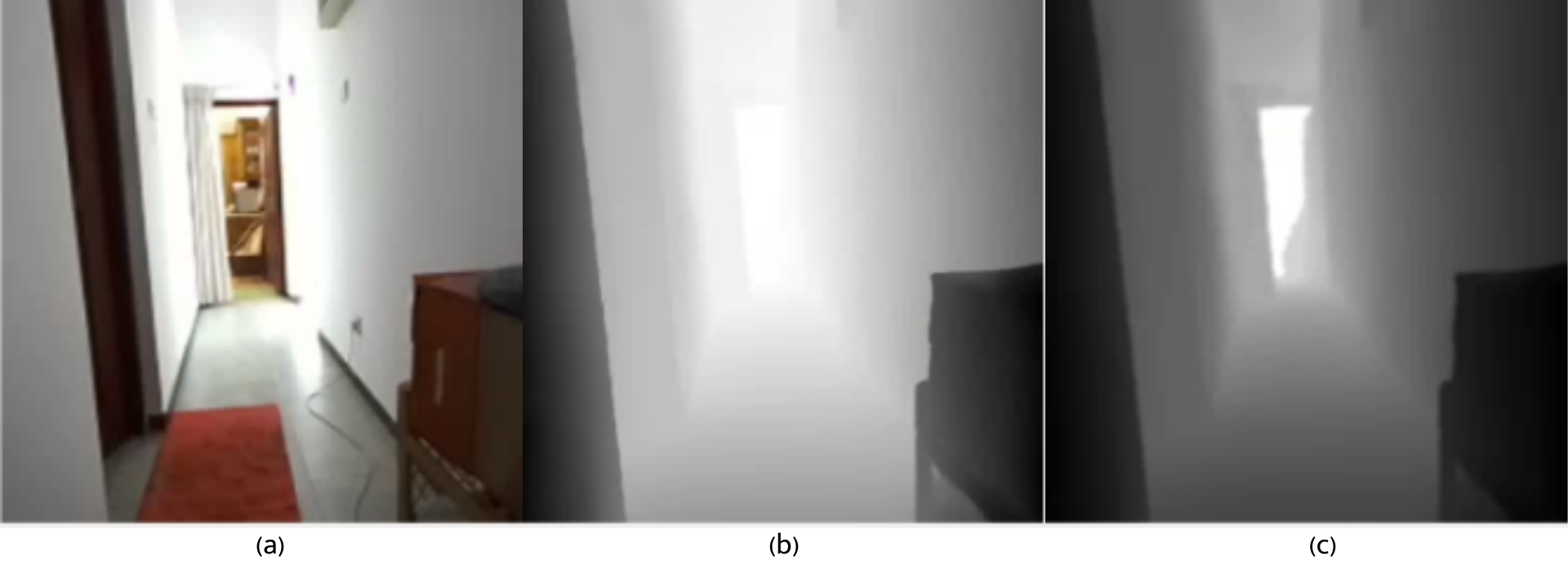}
      \caption{The original depth deceived our VLN system; the enhanced depth being closer to scenes in the simulation environment enabled successful navigation. (a) RGB image, (b) Depth image, (c) Gamma corrected depth image}
      \label{fig:enhancements}
   \end{figure}

\subsubsection{Data Collection and Fine Tuning}
\label{sec:datacollection}
For real-world fine-tuning and evaluation, data was collected in 6 environments consisting of 20 trajectories (2500+ RGB-D frames in total). These trajectories were deliberately designed to span residential, office, and university environments and to incorporate diverse navigation primitives such as corridor traversal, room entry, and landmark-based decision points. When combined with layout and lighting variations, this results in over 140 distinct navigation configurations.\\
The robot was manually controlled using a joystick, navigating through the predefined environments while capturing RGB images, depth data, and the corresponding natural language instructions. 
Augmentations were applied to account for the changes in lighting conditions such as brightness and contrast using empirically found scaling factors \(\alpha = 1.5\) and \(\beta = 1.2\) (for RGB) and Gamma correction with a gamma value of $0.35$ (for depth).
Through visual comparison with simulated input and evaluation through experimentation, we selected these values as the best fit for our hardware. This selection process served as a qualitative ablation study, where we iteratively adjusted parameters to minimize the perceptual domain gap between the real-world sensor streams and the Matterport3D simulation rendering. We observed that inputs lacking these specific photometric corrections resulted in feature embeddings that were unrecognizable to the pre-trained encoders, leading to immediate navigation failures regardless of the trajectory. Consequently, we established these values as the necessary baseline for effective sim-to-real transfer before proceeding to quantitative evaluation. \\As illustrated in Fig.~\ref{fig:enhancements}, we see a clear improvement of the resulting images after the corrections. These values worked well for all the 6 indoor environments.

\subsubsection{Evaluation in Real-World Environment}
\begin{figure*}[ht]
  \centering
  \includegraphics[width=1.0\linewidth]{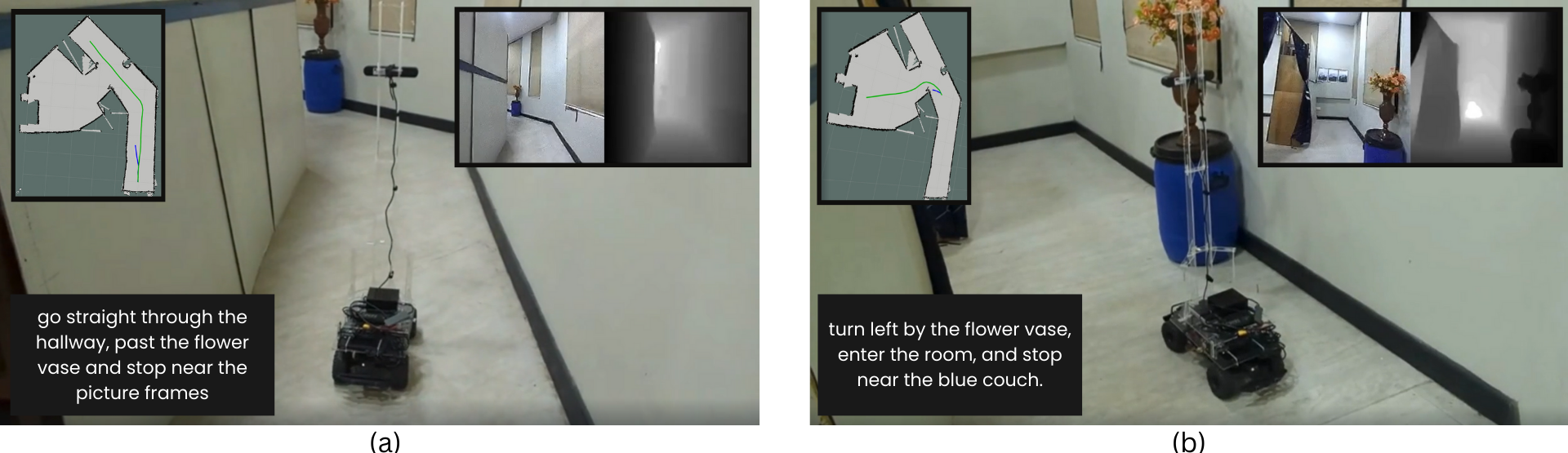}
  \caption {The robot navigation in the same environment with two different instructions for different routes (a), (b), including the robot's vision, instruction and path tracking.}  
  \label{fig:The robot navigation}
\end{figure*}

\begin{figure}[t]
  \centering
  \includegraphics[width=1.0\linewidth]{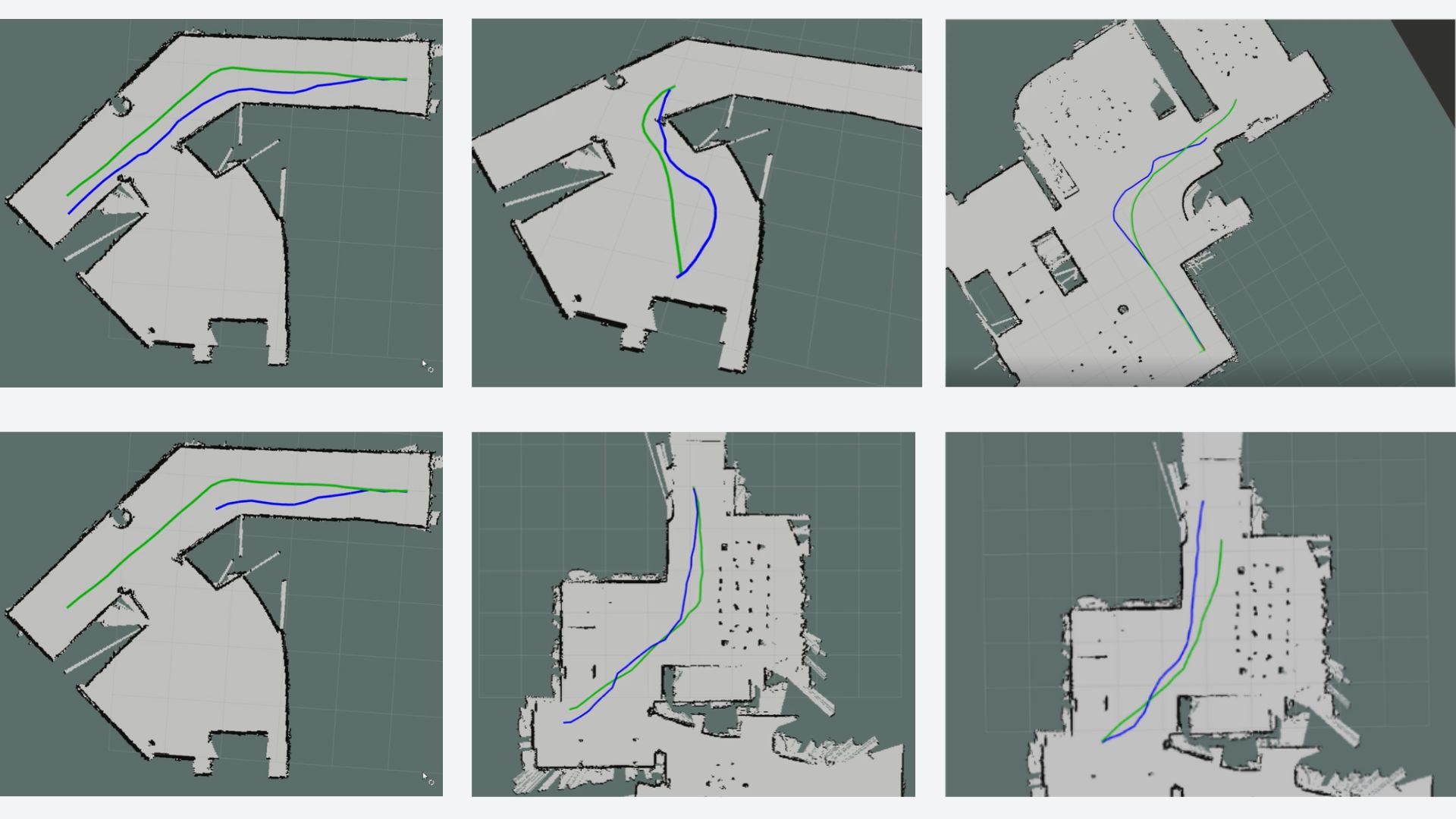}
   \caption{ Different routes tracked in several locations for different instructions. The green line represents the ideal path for a particular instruction, while the blue line represents the robot path for that instruction. }
  \label{fig:Path tracking}
\end{figure}

To evaluate the model, experiments were conducted using the mobile robot platform in 6 different indoor environments for 18 episodes, including offices, hallways, and open spaces, to test the model’s adaptability to various real-world conditions (Fig.~\ref{fig:The robot navigation}). The environments included a range of obstacles, lighting variations, and different surface types, reflecting the complexity of real-world navigation tasks. To validate the results of our system, we ran a parallel mapping and localization system to generate the real and actual paths as seen in Fig.~\ref{fig:Path tracking}. The real paths were generated by human navigation using the same hardware.

\subsection{Evaluation Metrics}
The performance of the VLN system was evaluated using two key metrics:
\subsubsection{Success weighted by Path Length (SPL) \cite{anderson2018evaluation}} This metric (Eq.~(\ref{eq:spl})) evaluates the success rate of reaching target destinations, adjusted by the efficiency of the path taken. It combines the completion of navigation tasks with the optimality of the paths, providing a comprehensive measure of navigation performance.
    \begin{equation}
        \mathrm{SPL} = \frac{1}{N} \sum_{i=1}^{N} S_i \frac{l_i}{\max(p_i, l_i)}
        \label{eq:spl}
    \end{equation}
where $N$: number of episodes, $S_i$: success of an episode, $p_i$: robot path length, $l_i$: ideal path length, and $i$: episode.\\
\subsubsection{Normalized Dynamic Time Warping (nDTW) \cite{anderson2018evaluation}} nDTW (Eq.~(\ref{eq:ndtw1})) was used to compare the robot's actual trajectory against a reference trajectory derived from natural language instructions. This metric quantifies the alignment between the predicted and ideal paths, offering insights into the accuracy and precision of the robot's navigation.
\begin{equation}
\label{eq:ndtw1}
\mathrm{nDTW} = 1 - \frac{{D}}{D_{\text{min}}(i_k,j_k)}
\end{equation}
where,
\begin{equation}
\label{eq:ndtw2}
D_{\text{min}}(i_k,j_k) = \min(D_{\text{min}}(i_{k-1},j_{k-1} ) + d(i_k,j_k|i_{k-1},j_{k-1}))
\end{equation}
\begin{equation}
D = \sum_{k} d(i_k,j_k).
\end{equation}
$D$: DTW distance (overall path cost), $d$: Euclidean distance between $(i_k,j_k)$ and  $(i_{k-1},j_{k-1})$, $(i_k,l_i)$: grid points in a warping sequence.

\subsection{Results}
\subsubsection{Performance in Simulation}
In the simulation environment, the VLN-CE model achieved an  SPL score of $0.32$ and an nDTW score of $0.52$ across validation episodes. These results indicate a considerable accuracy level as a baseline in following natural language instructions and navigating through the simulated environments. 
\subsubsection{Performance in Real-World Environments}
In real-world settings, the fine-tuned model achieved an average SPL score of $0.26$ and an nDTW score of $0.46$. While these scores are slightly lower than those observed in simulation, they represent a significant improvement over baseline models trained solely in simulation without fine-tuning; the model effectively adapted to real-world conditions, managing variations in lighting and sensor input while accurately interpreting and following natural language instructions.

\subsubsection{Analysis of Results}
 Table \ref{real_env_results} highlight the effectiveness of the fine-tuning process in bridging the sim-to-real gap in the real environments tested. Table \ref{comparison} summarizes these results in comparison with those of the simulation and other existing models. 
The performance drop from simulation to real (19\% for SPL and 12\% for nDTW) is due to real-world factors such as sensor noise (e.g., imperfect depth readings) and visual discrepancies (e.g., lighting conditions), and environmental dynamics which are not present in the simulation that can lead the robot taking longer, inferior paths or even failing to reach the goal in extreme cases. 
It directly affects the success rate ($S_i$) and path length ($p_i$) values in SPL score (Eq.~(\ref{eq:spl})). Moreover, longer and inferior robot paths can result in a decrease in nDTW (Eq.~(\ref{eq:ndtw1})) due to considerable deviation from ideal paths. 
Nevertheless, the model’s ability to maintain high SPL and nDTW scores demonstrates its robustness and adaptability to real-world challenges. The sim-to-real \cite{anderson2021sim} implementation shows improved SPL and nDTW values when provided with a map (Table~\ref{comparison}); however, by comparison of the 3rd and 4th rows (Table~\ref{comparison}), it is evident that our model surpasses its performance in scenarios where no map is available. Even though The feature fields implementation by Zihan Wang et al. \cite{wang2024sim} outperforms ours slightly in simulation environments due to the usage of panoramic VLN models, it drops in performance significantly in real environments compared to ours.

\begin{table}[t]
\caption{Model performance in real-world: 6 distinct environments} 
\label{real_env_results}
\begin{center}
\begin{tabular}{@{}lcccccr@{}}
\toprule
Environment no. & 1 & 2 & 3 & 4 & 5 & 6\\
\midrule
SPL value & 0.26 & 0.34 & 0.22 & 0.27 & 0.24 & 0.23\\
\midrule
nDTW value & 0.46 & 0.53 & 0.42 & 0.54 & 0.42 & 0.41\\
\bottomrule
\end{tabular}
\end{center}
\end{table}

\begin{table}[t]
\caption{Model performance in simulation \& real-world test scenarios in comparison with existing models}
\label{comparison}
\begin{center}
\begin{tabular}{@{}lccccr@{}}
\toprule
& Map & Setting & SPL $\uparrow$ & nDTW $\uparrow$\\
\midrule
VLN-CE \cite{krantz2020beyond} & \XSolidBrush & Simulation &0.32 & 0.52\\
\midrule
sim-to-real \cite{anderson2021sim}  & \Checkmark & Real & 0.44 & 0.39\\
\midrule
sim-to-real \cite{anderson2021sim} & \XSolidBrush & Real & 0.22 &  0.30\\
\midrule
Feature Fields \cite{wang2024sim} & \Checkmark & Simulation & 0.29 &  -\\
\midrule
Feature Fields \cite{wang2024sim} & \Checkmark & Real & 0.18 &  -\\
\midrule
\textbf{Ours} & \XSolidBrush & Real & \textbf{0.26} & \textbf{0.46}\\
\bottomrule
\end{tabular}
\end{center}
\end{table}


\section{CONCLUSION}
In this work, we implement the first sim-to-real transfer of a Vision-and-Language Navigation in Continuous Environment (VLN-CE) model to a real-world setting using an Ackermann-steered robot. Our approach, leveraging a cross-modal attention model trained on the VLN-CE dataset, effectively navigates unseen indoor environments without relying on navigation graphs or 360-degree views. With targeted fine-tuning using real-world data and minimal photometric adjustments, we achieve performance in real-world settings with a minimal performance drop (from simulation to real) of approximately 19\% for SPL and 12\% for nDTW. This can be attributed to real-world factors such as sensor noise, lighting variations, and environmental dynamics.

While our results are promising, future work should focus on further narrowing the sim-to-real gap by enhancing the model’s robustness through improved transfer techniques and training on more diverse simulated environments. Given our use of an Ackermann-steered platform, a natural progression is the extension of VLN to highly dynamic outdoor environments, such as those encountered by autonomous vehicles. This transition presents unique challenges, including the need for enhanced camera dynamic range and more photometric adjustments-specifically adaptive scaling factors and gamma corrections-to handle varying outdoor lighting conditions. Future research will also focus on scaling our training set with outdoor-specific trajectories and a richer linguistic vocabulary to accommodate the complexity of outdoor instructions. In the initial stages of outdoor deployment, we will prioritize safety through the integration of immediate manual override controls and real-time obstacle avoidance protocols.

\bibliographystyle{IEEEtran}
\bibliography{references}

@inproceedings{li2023kerm,
  title={Kerm: Knowledge enhanced reasoning for vision-and-language navigation},
  author={Li, Xiangyang and Wang, Zihan and Yang, Jiahao and Wang, Yaowei and Jiang, Shuqiang},
  booktitle={Proceedings of the IEEE/CVF Conference on Computer Vision and Pattern Recognition},
  pages={2583--2592},
  year={2023}
}

@inproceedings{chen2022think,
  title={Think global, act local: Dual-scale graph transformer for vision-and-language navigation},
  author={Chen, Shizhe and Guhur, Pierre-Louis and Tapaswi, Makarand and Schmid, Cordelia and Laptev, Ivan},
  booktitle={Proceedings of the IEEE/CVF Conference on Computer Vision and Pattern Recognition},
  pages={16537--16547},
  year={2022}
}

@inproceedings{shah2023lm,
  title={Lm-nav: Robotic navigation with large pre-trained models of language, vision, and action},
  author={Shah, Dhruv and Osi{\'n}ski, B{\l}a{\.z}ej and Levine, Sergey and others},
  booktitle={Conference on robot learning},
  pages={492--504},
  year={2023},
  organization={PMLR}
}

@inproceedings{krantz2021waypoint,
  title={Waypoint models for instruction-guided navigation in continuous environments},
  author={Krantz, Jacob and Gokaslan, Aaron and Batra, Dhruv and Lee, Stefan and Maksymets, Oleksandr},
  booktitle={Proceedings of the IEEE/CVF International Conference on Computer Vision},
  pages={15162--15171},
  year={2021}
}

@inproceedings{anderson2018vision,
  title={Vision-and-language navigation: Interpreting visually-grounded navigation instructions in real environments},
  author={Anderson, Peter and Wu, Qi and Teney, Damien and Bruce, Jake and Johnson, Mark and S{\"u}nderhauf, Niko and Reid, Ian and Gould, Stephen and Van Den Hengel, Anton},
  booktitle={Proceedings of the IEEE conference on computer vision and pattern recognition},
  pages={3674--3683},
  year={2018}
}

@inproceedings{chen2021topological,
  title={Topological planning with transformers for vision-and-language navigation},
  author={Chen, Kevin and Chen, Junshen K and Chuang, Jo and V{\'a}zquez, Marynel and Savarese, Silvio},
  booktitle={Proceedings of the IEEE/CVF Conference on Computer Vision and Pattern Recognition},
  pages={11276--11286},
  year={2021}
}

@inproceedings{krantz2020beyond,
  title={Beyond the nav-graph: Vision-and-language navigation in continuous environments},
  author={Krantz, Jacob and Wijmans, Erik and Majumdar, Arjun and Batra, Dhruv and Lee, Stefan},
  booktitle={Computer Vision--ECCV 2020: 16th European Conference, Glasgow, UK, August 23--28, 2020, Proceedings, Part XXVIII 16},
  pages={104--120},
  year={2020},
  organization={Springer}
}

@inproceedings{irshad2021hierarchical,
  title={Hierarchical cross-modal agent for robotics vision-and-language navigation},
  author={Irshad, Muhammad Zubair and Ma, Chih-Yao and Kira, Zsolt},
  booktitle={2021 IEEE International Conference on Robotics and Automation (ICRA)},
  pages={13238--13246},
  year={2021}
}

@inproceedings{savva2019habitat,
  title={Habitat: A platform for embodied ai research},
  author={Savva, Manolis and Kadian, Abhishek and Maksymets, Oleksandr and Zhao, Yili and Wijmans, Erik and Jain, Bhavana and Straub, Julian and Liu, Jia and Koltun, Vladlen and Malik, Jitendra and others},
  booktitle={Proceedings of the IEEE/CVF international conference on computer vision},
  pages={9339--9347},
  year={2019}
}

@article{anderson2018evaluation,
  title={On evaluation of embodied navigation agents},
  author={Anderson, Peter and Chang, Angel and Chaplot, Devendra Singh and Dosovitskiy, Alexey and Gupta, Saurabh and Koltun, Vladlen and Kosecka, Jana and Malik, Jitendra and Mottaghi, Roozbeh and Savva, Manolis and others},
  journal={arXiv preprint arXiv:1807.06757},
  year={2018}
}

@inproceedings{deng2009imagenet,
  title={Imagenet: A large-scale hierarchical image database},
  author={Deng, Jia and Dong, Wei and Socher, Richard and Li, Li-Jia and Li, Kai and Fei-Fei, Li},
  booktitle={IEEE conference on computer vision and pattern recognition},
  pages={248--255},
  year={2009}
}

@article{ramakrishnan2021habitat,
  title={Habitat-matterport 3d dataset (hm3d): 1000 large-scale 3d environments for embodied ai},
  author={Ramakrishnan, Santhosh K and Gokaslan, Aaron and Wijmans, Erik and Maksymets, Oleksandr and Clegg, Alex and Turner, John and Undersander, Eric and Galuba, Wojciech and Westbury, Andrew and Chang, Angel X and others},
  journal={arXiv preprint arXiv:2109.08238},
  year={2021}
}

@inproceedings{anderson2021sim,
  title={Sim-to-real transfer for vision-and-language navigation},
  author={Anderson, Peter and Shrivastava, Ayush and Truong, Joanne and Majumdar, Arjun and Parikh, Devi and Batra, Dhruv and Lee, Stefan},
  booktitle={Conference on Robot Learning},
  pages={671--681},
  year={2021}
}

@inproceedings{vln-pano2real,
  title={Sim-to-Real Transfer for Vision-and-Language Navigation},
  author={Peter Anderson and Ayush Shrivastava and Joanne Truong and Arjun Majumdar and Devi Parikh and Dhruv Batra and Stefan Lee},
  booktitle={Conference on Robot Learning (CoRL)},
  year={2020}
}

@inproceedings{pennington2014glove,
  title={Glove: Global vectors for word representation},
  author={Pennington, Jeffrey and Socher, Richard and Manning, Christopher D},
  booktitle={Proceedings of the 2014 conference on empirical methods in natural language processing (EMNLP)},
  pages={1532--1543},
  year={2014}
}

@inproceedings{he2016deep,
  title={Deep residual learning for image recognition},
  author={He, Kaiming and Zhang, Xiangyu and Ren, Shaoqing and Sun, Jian},
  booktitle={Proceedings of the IEEE conference on computer vision and pattern recognition},
  pages={770--778},
  year={2016}
}

@inproceedings{wijmans2019dd,
  title={Dd-ppo: Learning near-perfect pointgoal navigators from 2.5 billion frames},
  author={Wijmans, Erik and Kadian, Abhishek and Morcos, Ari and Lee, Stefan and Essa, Irfan and Parikh, Devi and Savva, Manolis and Batra, Dhruv},
  booktitle={Proceedings of the International Conference on Learning Representations (ICLR)},
  year={2019}
}

@inproceedings{hermann2020learning,
  title={Learning to follow directions in street view},
  author={Hermann, Karl Moritz and Malinowski, Mateusz and Mirowski, Piotr and Banki-Horvath, Andras and Anderson, Keith and Hadsell, Raia},
  booktitle={Proceedings of the AAAI Conference on Artificial Intelligence},
  volume={34},
  number={07},
  pages={11773--11781},
  year={2020}
}

@article{an2024etpnav,
  title={Etpnav: Evolving topological planning for vision-language navigation in continuous environments},
  author={An, Dong and Wang, Hanqing and Wang, Wenguan and Wang, Zun and Huang, Yan and He, Keji and Wang, Liang},
  journal={IEEE Transactions on Pattern Analysis and Machine Intelligence},
    pages = {1--16},
  year={2024}
}

@inproceedings{an2023bevbert,
  title={Bevbert: Multimodal map pre-training for language-guided navigation},
  author={An, Dong and Qi, Yuankai and Li, Yangguang and Huang, Yan and Wang, Liang and Tan, Tieniu and Shao, Jing},
  booktitle={Proceedings of the IEEE/CVF International Conference on Computer Vision},
  pages={2737--2748},
  year={2023}
}

@article{daniuvsis2021topological,
  title={Topological navigation graph framework},
  author={Daniu{\v{s}}is, Povilas and Juneja, Shubham and Valatka, Lukas and Petkevi{\v{c}}ius, Linas},
  journal={Autonomous Robots},
  volume={45},
  number={5},
  pages={633--646},
  year={2021}
}

@article{cho2014learning,
  title={Learning phrase representations using RNN encoder-decoder for statistical machine translation},
  author={Cho, Kyunghyun},
  journal={arXiv preprint arXiv:1406.1078},
  year={2014}
}

@inproceedings{Quigley2009ROSAO,
  title={ROS: an open-source Robot Operating System},
  author={Morgan Quigley},
  booktitle={IEEE International Conference on Robotics and Automation},
  year={2009}
}

@article{DurrantWhyte2006SimultaneousLA,
  title={Simultaneous localization and mapping: part I},
  author={Hugh F. Durrant-Whyte and Tim Bailey},
  journal={IEEE Robotics \& Automation Magazine},
  year={2006},
  volume={13},
  pages={99--110}
}

@inproceedings{zhu2021soon,
  title={Soon: Scenario oriented object navigation with graph-based exploration},
  author={Zhu, Fengda and Liang, Xiwen and Zhu, Yi and Yu, Qizhi and Chang, Xiaojun and Liang, Xiaodan},
  booktitle={Proceedings of the IEEE/CVF Conference on Computer Vision and Pattern Recognition},
  pages={12689--12699},
  year={2021}
}

@inproceedings{qi2020reverie,
  title={Reverie: Remote embodied visual referring expression in real indoor environments},
  author={Qi, Yuankai and Wu, Qi and Anderson, Peter and Wang, Xin and Wang, William Yang and Shen, Chunhua and Hengel, Anton van den},
  booktitle={Proceedings of the IEEE/CVF Conference on Computer Vision and Pattern Recognition},
  pages={9982--9991},
  year={2020}
}

@article{graves2005framewise,
  title={Framewise phoneme classification with bidirectional LSTM and other neural network architectures},
  author={Graves, Alex and Schmidhuber, J{\"u}rgen},
  journal={Neural networks},
  volume={18},
  number={5-6},
  pages={602--610},
  year={2005},
  publisher={Elsevier}
}

@article{wang2024sim,
  title={Sim-to-real transfer via 3d feature fields for vision-and-language navigation},
  author={Wang, Zihan and Li, Xiangyang and Yang, Jiahao and Liu, Yeqi and Jiang, Shuqiang},
  journal={arXiv preprint arXiv:2406.09798},
  year={2024}
}

@inproceedings{dias2020deep,
  title={Deep learning of augmented reality based human interactions for automating a robot team},
  author={Dias, Adhitha and Wellaboda, Hasitha and Rasanka, Yasod and Munasinghe, Menusha and Rodrigo, Ranga and Jayasekara, Peshala},
  booktitle={2020 6th International Conference on Control, Automation and Robotics (ICCAR)},
  pages={175--182},
  year={2020},
  organization={IEEE}
}

%
%
%
%

\end{document}